\documentclass[letterpaper]{article} 
\usepackage{aaai25}  
\usepackage{times}  
\usepackage{helvet}  
\usepackage{courier}  
\usepackage[hyphens]{url}  
\usepackage{graphicx} 
\usepackage{natbib}  
\usepackage{caption} 
\usepackage{algorithm}
\usepackage{algorithmic}

\usepackage{newfloat}
\usepackage{listings}

\usepackage{amsmath}
\usepackage{booktabs}
\usepackage{multirow}
\usepackage{array}
\usepackage{subfigure}
\usepackage{hyperref}

\DeclareCaptionStyle{ruled}{labelfont=normalfont,labelsep=colon,strut=off} 
\floatstyle{ruled}
\newfloat{listing}{tb}{lst}{}
\floatname{listing}{Listing}

\title{Strategies for Improving NL-to-FOL Translation with  LLMs: \\Data Generation, Incremental Fine-Tuning, and Verification}
\author {
    Ramya Keerthy Thatikonda\textsuperscript{\rm 1},
    Jiuzhou Han\textsuperscript{\rm 1},
    Wray Buntine\textsuperscript{\rm 2},
    Ehsan Shareghi\textsuperscript{\rm 1}
}
\affiliations {
    \textsuperscript{\rm 1}Department of Data Science \& AI, Monash University\\
    \textsuperscript{\rm 2}College of Engineering and Computer Science, VinUniversity\\
    ramya.thatikonda1@monash.edu
}

\begin{document}
\nocopyright 
\maketitle

\begin{abstract}
\label{sc: Abstract}
Logical reasoning is a fundamental task in natural language processing that presents significant challenges to Large Language Models (LLMs). The inherent characteristics of logical reasoning makes it well-suited for symbolic representations such as first-order logic (FOL). Research in symbolic logical reasoning explored FOL generation using state-of-the-art LLMs (i.e., GPT-4) to produce FOL translations of natural language (NL) statements, but errors in translation are usually not the focus. We address this by categorizing the translation errors in FOL statements generated by LLMs, specifically for deductive logical reasoning tasks. In order to make progress towards improving the quality of FOL translations for smaller language models such as LLaMA-2 13B and Mistral 7B, we create \textsc{ProofFOL}, a high-quality FOL-annotated subset of ProofWriter dataset using GPT-4o. The models fine-tuned on this silver standard data achieve a significant gain in performance when compared to larger language models such as LLaMA-2 70B. In addition to improving the model using large data, we also tackle the issue of data scarcity and introduce an incremental framework encompassing of data augmentation and verification steps. In the augmentation process, a single pair of (premises, conclusion) is split into multiple new instances based on the predicates and FOLs. This data is used for fine-tuning, and the inference on this model generates FOLs with fewer errors over the model trained on the original data. Our investigation on the translation errors leads to generation of a perturbation dataset consisting of simulated NL-to-FOL translation errors and their corresponding corrections. This data is used to train a verifier, which corrects potential syntactic and semantic FOL translation errors. We demonstrate an efficient method for making the most of a limited existing human-annotated dataset. Our results show state-of-the-art performance for ProofWriter and ProntoQA datasets using \textsc{ProofFOL} on LLaMA-2 and Mistral models. \footnote{The code for fine-tuning, augmentation and verification, and \textsc{ProofFOL} dataset are available at https://github.com/RamyaKeerthy/Translation-NL2FOL.}
\end{abstract}

\section{Introduction}
Recent state-of-the-art methods for logical reasoning from natural language (NL) descriptions work via translation \cite{pan2023logic,ye2024satlm,olausson2023linc}. An LLM is tasked to translate statements from NL to first order logic (FOL), which is then sent for execution to external SMT solvers such as Z3~\cite{de2008z3} and Prover9~\cite{mccune2005release}.
Recent work \cite{yang2023harnessing} has highlighted the systematic errors that even the most capable LLMs (such as GPT-4) make during translation of a \emph{single} NL statement into its corresponding FOL.  A realistic logical reasoning scenario is much more demanding as it involves several premise statements followed by a conclusion to be verified. This requires consistency of NL-to-FOL translations (e.g., in predicate naming or translation of logical operators) across several statements. However, very little is explored on the pattern of syntactic and semantic errors LLMs make during such translation scenario.  

Existing approaches to reduce NL-to-FOL translation errors are of limited impact.  They rely on the LLM's capability to understand and self-correct the translation inaccuracies only based on the error message from the external SMT solver~\cite{pan2023logic}, but the ability to comprehend an error message is often restricted to larger scale of models (i.e., 175B), and not possible in smaller LLMs~(i.e., 7B, 13B). Also, while syntactic translation mistakes (e.g., \emph{misplaced operator: $P\wedge\rightarrow Q$}, or \emph{missing quantifiers: $P(x)\rightarrow Q(x)$}) result in run-time errors by tools, many semantic errors (e.g., \emph{use of incorrect quantifiers: All men are mortal. $\exists x (\text{Man}(x) \rightarrow \text{Mortal}(x))$
}) are passed through SMT tools without causing any run-time error, limiting the ability of such an approach to correct less trivial errors. A straightforward solution to reduce errors is to fine-tune the smaller models on large scale of NL-to-FOL translation data. However, the existing datasets offer very little support, with FOLIO~\cite{han2022folio} as the only human-annotated dataset to have around $1k$ examples of NL (premises, conclusion) and their FOL translations. \textsc{Malls}~\cite{yang2023harnessing} is another synthetic dataset which has $28k$ pair of single statements and FOL translations. Even in the presence of larger scale fine-tuning data, it is still desired to have a correction mechanism for smaller models that could catch both syntactic and semantic errors on-the-fly at inference time.

In this paper we combine a number of methods to reduce NL-to-FOL translation errors. First, to overcome the shortage of ground truth for fine-tuning, we utilise GPT-4o to create \textsc{ProofFOL}, a new large corpus of (premises, conclusion) pairs and corresponding FOL translations. More concretely, we use the existing pairs of (premises, conclusion) of ProofWriter dataset~\cite{tafjord2020proofwriter} and prompt GPT-4 to generate the corresponding FOL translations. To ensure the correctness of the FOL translations, we pass them through Prover9 and filter out examples that produce outputs (i.e., \emph{true, false, uncertain}) that do not match the ground truth in ProofWriter. \textsc{ProofFOL} is the largest existing dataset consisting of in $10424$ examples of (premises, conclusion and their corresponding FOL translations. We show that fine-tuned LLaMA-2 13B \cite{touvron2023llama} and Mistral 7B \cite{jiang2023mistral} models on \textsc{ProofFOL} outperform larger baselines such as LLaMA-2 70B and Mixtral $8\times7\text{B}$, both in terms of translation quality and performance on logical reasoning task on ProofWriter and ProntoQA~\cite{prontoqa}. 

Second, to effectively utilize existing scarce but high-quality human-generated data (i.e., FOLIO), we propose a set of incremental techniques for both the training and inference phases. More concretely, to increase the number of training instances, we split each record of FOLIO into multiple records for an incremental training process where the model is trained to first produce the predicates of premises and conclusions, and then to produce the FOL of each statements of premises and conclusion one by one. This model further improves the predictive accuracy on FOLIO by $41\%$.

Third and finally, to allow for fine-grained correction of semantic and syntactic errors at inference time, we train separate T5~\cite{raffel2020exploring} models to exclusively verify predicates and FOLs on-the-fly, and apply necessary corrections when needed. In particular, we first define categories of semantic and syntactic errors made by LLMs during NL-to-FOL Translation. To simulate these errors, we  apply controlled perturbations on ground truth labels of FOLIO and create a new training set consisting of statements and their corresponding \emph{perturbed} predicates and FOLs. Then, we train T5 models which take as input statements and their predicates, or statements and their FOLs, and either verifies their correctness (i.e., outputs \emph{correct}) or outputs corrected versions. This added mechanism brought a further $17\%$ improvement on FOLIO.

 Our findings highlight that data is crucial for surpassing the current performance limits of several large language models, particularly when employing more accessible models for logical reasoning tasks. Our data generation pipeline allows us to create, \textsc{ProofFOL}, the largest FOL-annotated logical reasoning dataset to this date. Our incremental training presents an effective data augmentation method particularly useful for data scarce conditions, while the verifier mechanism and corresponding training protocol offers a promising pathway to verify correctness of symbolic forms generated by LLMs at inference time.

\section{NL-to-FOL Translation Errors}
\label{sec:nlfol}
First-order logic (FOL) is a form of logic representation that covers the use of variables, functions, and quantifiers. FOLs are suitable for logical reasoning tasks involving natural language. Large Language Models (LLMs) are shown to be capable of translating natural language into various types of formal representations with varying degree of success. 
Among these formalism, NL-to-FOL translation presents unique syntactic (FOL syntax) and semantic (interpreting the meaning of the NL statements) challenges. We first attempt to categorize these syntactic and semantic errors in this section. This will serve as the basis for our data perturbation protocol to train FOL verifiers (presented shortly).

The translation of language to FOL follows grammatical rules and any deviation from the rules can cause syntax errors. For the rule \textit{``every free variable assigned to a predicate has a quantifier''} can be applied to the statement ``Green people are blue'' with predicates `green' and `blue', a free variable `$x$' and a quantifier `$\forall$' to quantify the variable. A missing quantifier can be a parsing error in syntax for this rule. The Prover9 tool, used as a logical solver in our paper, is designed to send a feedback when encountered with syntax errors. We use this to detect and categorize the syntactic errors during translation into the following categories: 
\begin{itemize}
    \item \textit{Parsing errors:} This occurs when the logical statement has missing or invalid operators, or missing parentheses. 
    \item \textit{Type errors:} The tool returns \emph{none} value when a sentence level discrepancy, e.g. missing quantifier, is encountered. 
    \item \textit{Token errors}: This type of error is observed when an invalid token is part of the FOL, such as a \$ symbol.
\end{itemize}
 
Semantic errors are generally not detectable as they follow the structure of formal language, but can be identified by observing the incorrect predictions of the parsable FOL statements. This process is tedious as it requires effort to scan through translations and point the occurrence of the error. An example of semantic error is when the language model predicts a wrong operator. In case of the sentence ``All rabbits have fur'', the FOL \(\exists{x}(\text{Rabbit(x)} \rightarrow \text{Have(x, fur)})\) incorrectly represents the quantifier of the sentence. Semantic errors are broadly categorized as
\begin{itemize}
    \item \textit{Sense errors:} This is a general form of semantic error where the prediction from the tool is incorrect, pointing to inaccuracy between natural language and FOL.
    \item \textit{Arities errors:} This is a specific error thrown by a tool to indicate that there are duplicate predicate values with mismatch in number of arguments. 
\end{itemize}

These errors are further studied manually to divide into more specific categories. Due to space constraints, we define them in detail in the Appendix. We provide an empirical distribution of these errors in Section~\ref{sec:main-results} and Figure~\ref{fig:sanki}.

\section{Incremental Fine-Tuning and Verification}
\subsection{Data Generation and Fine-tuning}
The alignment of a language model to follow instructions for a specific task can be accomplished by fine-tuning on substantial data. The task of formal translations require first-order logic of their natural language counterparts. Ideally, this task is at a passage level rather than sentence level, which makes it challenging for a language model to follow a required format. To overcome this, we need sufficient passage level translations, which are time-consuming to generate through human annotations. We introduce a streamlined process for generating this FOL data and ensuring correctness of the format, grammar, and order of the translations.

\paragraph{Data Generation}
For the data generation process, we pick ProofWriter which comes with large number of training records, each consisting of multipe premises and a conclusion, and variations in depth of reasoning. The format of the FOLs is set to be consistent with the ``Prover9" theorem prover, which has a human understandable, formal language syntax.\footnote{Given the requisite for diversity in syntax and semantic, we first chose a few combinations of demonstrations, and ran a small scale experiment through GPT-4o for each combination. We then selected the most optimal demonstrations with the least format and translation issues in the resulting generated FOL data. We report these final demonstrations in Appendix.} The fixed format of demonstrations is adapted from \citet{pan2023logic}, where we generate predicates followed by first-order logic statements for each sentence. We change the formal language to Prover9 and keep it fixed for all datasets. 

At a glance, the output from the GPT model has formatting issues, such as assigning numbers to each generation, explaining the task before generation, and solving for conclusion after producing translations. These issues are parsed using pattern matching to obtain the maximum number of translations. After the parsing stage, the syntax check is done by the Prover9, where the tool can provide unique feedback for each form of syntax error. When analysing these errors, we observed grammar rules that can be fixed in the tool to include unicode decoding, allow unordered quantifiers and support negation of a full formula. This addition of grammar rules minimized the penalization for translations. The semantic errors were identified by comparing the ground truth label from ProofWriter with Prover9-generated output based on FOLs. This systematic pipeline enabled the retention of 70\% of the silver-standard data. We term this dataset \textsc{ProofFOL}, which consists of $10424$ pairs of (Premises, Conclusions), deduction label, and corresponding FOLs.    

\paragraph{Supervised Fine-tuning (SFT)}
Each input $x$ to the SFT is a set of premise statements $P_x (=\{P_1, P_2, \ldots, P_n\})$, a conclusion $C_x$, and an instruction ($I$) represented as \([P_x, C_x, I]\). In order to avoid over-fitting the model to certain spurious patterns in GPT-4o translations, we include human-generated dataset (FOLIO) in the mix. We create a set of models built on the full dataset, providing a perspective for model behaviour with size of the dataset. Since this SFT model employs both existing and synthetic data, it imposes a dilemma of the effect of gold-standard data on the results when compared to the generated silver-standard. To validate the reliability of our generated data, we fine-tune the model on a subset of the dataset and examine how increasing the data size impacts logical reasoning tasks. For a given input $x$, the output of SFT models are predicates of $x$ (denoted as $Pred_x$), and FOLs of its premises and conclusion, \([Pred_x, {FOL_{P_{x}}}, {FOL_{C_{x}}}]\). For experiments, see Section~\ref{sec:main-results}. 

\subsection{Incremental Techniques}
The SFT method described in the previous section uses the FOLIO dataset as just a small component of the overall approach. With high number of records associated with \textsc{ProofFOL}, we can assume that the in-distribution dataset will show a major improvement when compared to FOLIO. To enhance the performance of FOLIO dataset (and in general any similar data-scarce scenario), we introduce \emph{`Incremental Techniques'} for maximizing the use of limited data. These techniques encompass data augmentation, incremental fine-tuning and inference, and incremental verification of predicates and FOLs, creating a comprehensive setup for FOL generation. We further expand this method to a smaller subset of \textsc{ProofFOL} to simulate data scarce environment.

\subsubsection{Data Augmentation}
Supervised Fine-tuning a decoder model is technically an unannotated form of training as the \emph{supervised} part of the fine-tuning refers to the label that is passed with the input. During a vanilla SFT process, we pass the whole output along with the input, and the model performs inference as a text completion task. This motivates our data augmentation method, where the model examines smaller part of the output rather than training on the whole output at a time. The FOL translation is one such task where the output sequence is lengthy and the model can deviate from generating the correct syntax. 

To initiate the data augmentation process, we split the output of the original record to represent incremental data, where the first output is \([Pred_x]\), the second output is \([Pred_x, {FOL_{P_{1}}}]\), and so on till we reach the full output \([Pred_x, {FOL_{P_{1}}}...,{FOL_{P_{n}}}, {FOL_{C_{x}}}]\). This splits a single record into \(n+2\) records, where $n$ is the number of premises and `$+2$' is for predicate and conclusion generation. The input remains the same for all the records. This data augmentation increases the dataset size to about $7\times$ and $20\times$ for FOLIO and ProofWriter, respectively. With this enhanced data, we train a set of SFT models for FOL translation tasks.

\subsubsection{Inference}
The SFT for augmented data is performed using two instructions, indicating two types of tasks. The first instruction is \emph{``Generate predicates for the given natural language sentences."} to generate the predicates, and the second is \emph{``Given a premise and conclusion, generate the first order logic form of the premises and conclusion."} to generate the FOL statements. At inference, we provide the model with the instruction to generate the predicates. These predicates, along with the input, are then passed to the model to infer the FOL statements. We categorize FOL inference into two forms. 
\begin{itemize}
    \item \textbf{Vanilla Inference:} In vanilla inference, the model is provided with the natural language statements and is asked to generate the predicates and the FOL translations for the complete input.  
    \item \textbf{Incremental Inference:} In incremental inference, we first generate the predicates, and then limit the generation to a single FOL translation at a time by setting a low \texttt{maximum\_new\_token} parameter. For every FOL generation, we pass the previously generated values as input. For example, if we are generating \({FOL_{P_{3}}}\), the input would be \([P_x, C_x, Pred_x, {FOL_{P_{1}}}, {FOL_{P_{2}}}] \). 
\end{itemize}

\begin{figure*}[t]
  \centering
  \includegraphics[width=\textwidth]{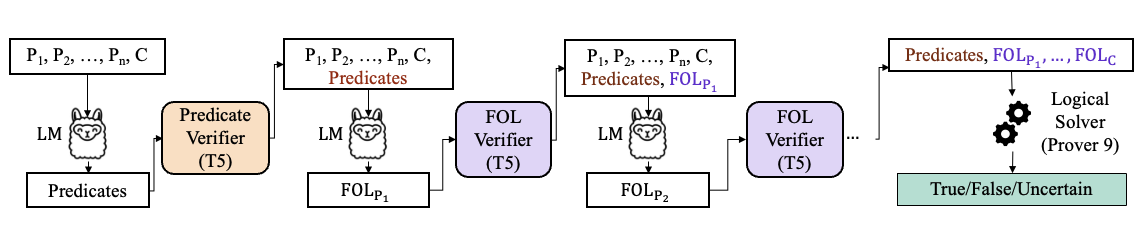}
  \caption{The overall flow of the incremental generation and verification at inference time. Here $\text{P}_i$s and $\text{C}$ denote Premises and Conclusion. For predicate and FOL generation, the output is run through the corresponding verifier. }
  \label{fig:datagen}
\end{figure*}

\subsubsection{Verification and Correction}
Since the incremental inference allows individual predicate and FOL generation, we train two verifiers; predicate and FOL, to detect and correct the potential errors. The term `verifier' refers to both the verification and correction processes. 

\begin{itemize}
    \item {\textbf{Predicate Verifier} The Predicate verifier takes the \([P_x, C_x, Pred_x]\), as the input, and evaluates the predicted predicate $Pred_x$. If the verifier considers $Pred_x$ correct, the output is just ‘correct’, otherwise the verifier will generate the corrected predicates set. To achieve this outcome, the verifier is trained on the perturbed predicates of the training dataset used for SFT. The perturbations for predicates are created based on the semantic errors related to predicates. Few perturbations used are; omitting predicate, omitting variable, adding variable, and adding duplicate predicate. The predicate perturbations are explained in detail in the Appendix.}

    \item {\textbf{FOL Verifier} The FOL verifier takes \([Pred_{\text{NL}}, FOL_{\text{NL}}, \text{NL}]\) as the input, where $FOL_{\text{NL}}$ represents predicted FOL form of a single premise ($P_i$) \emph{or} conclusion ($C_x$) statement (represented as $\text{NL}$). The verifier evaluates $FOL_{\text{NL}}$ as ‘correct’ or will generate a corrected FOL. Similar to predicate verifier, we use the original SFT training set and applied perturbations to train this Verifier model. The perturbations used in this method are crafted based on the common syntax and semantic errors in translations (Section~\ref{sec:nlfol}). Few perturbation used in this verifier are; changing quantifier position, omitting a quantifier, omitting parenthesis, tweaking negations, and replacing operators. We discuss these in detail in the Appendix.}
\end{itemize}

In addition to these perturbation instances, we incorporate real errors from the SFT training data. Specifically, we run the inference of the training data on the SFT model and collect the predicate and FOL, which do not match the ground truth, as errors. We add these errors to the fully-controlled created perturbations to form the `incorrect verifier training instances'. Finally, in order to verify `correct' predictions, we take the predicate and FOL values from the ground-truth and label them as ‘correct’. We generate the perturbation dataset from a seed of $1k$ examples of FOLIO and similarly for ProofWriter. For detail statistics on the size of the resulting perturbation datasets, please see Appendix.

Inspired by \citet{pive}, after the creation of the training data for the two verifiers, we fine-tune a T5-Large \cite{raffel2020exploring} model to work as the verifier. We train two separate models for the Predicate and FOL verification on each dataset. During incremental inference, the verifier is incorporated into the decoding phase, as shown in Fig \ref{fig:datagen}.

\section{Experimental Setup}
\subsection{\textsc{ProofFOL} Data Generation Specifications}
For the data generation process, we used the training set of ProofWriter Open World Assumption (OWA) with the highest complexity, depth of 5. The training data is passed along with 2-shot ProofWriter examples, which cover all the required operators, to \texttt{gpt4-o}. The format of FOL generation is adapted from \citet{pan2023logic}, where we generate predicates and FOL statements, each paired with a corresponding natural language text. We generate FOLs for 15000 training data points. These records are parsed using regex pattern match, specifically to separate the FOLs from unintended text. The predicted FOL and input NL statements are mapped and generations with incomplete or excess text are filtered out. The FOL statements are then passed to the Prover9 tool to perform deductive reasoning, where the tool returns `True', `False', `Unknown', or `None' in case of an error. The generated labels are compared against the ground truth and any mismatch is removed. The prompt templates and statistics of \textsc{ProofFOL} are provided in the Appendix. 

\subsection{Vanilla SFT}
We choose two families of models; LLaMA-2 and Mistral, and examine their performance on three deductive logical benchmarks; ProofWriter, FOLIO, and ProntoQA. The output of the LLMs are either based on the truth value of the conclusion given the NL format of premises and conclusions (denoted in results as Standard), or the formal language translations of the input which is then passed to Prover9 for generating the output (denoted in results as FOL). The initial baselines use in-context learning (ICL) for standard and FOL generation. The few-shot examples for standard generation are randomly sampled from the training data, ensuring a balanced output distribution. Since ProntoQA is a test-set only benchmark, we use ProofWriter demonstrations during inference on ProntoQA. We consider ProntoQA as out-of-distribution (OOD) since it was not represented in the SFT data. The few-shot FOL generation is similar to the process used in data generation, with variations in examples. The number of few-shot examples and their respective templates are presented in the Appendix. 

SFT with \textsc{ProofFOL} is performed on two models; LLaMA-2 13B and Mistral 7B. These models are selected based on their low computational cost and model parameters. The models are fine-tuned for 3-epochs with a fixed batch size and LoRA~\cite{hu2022lora}. The inference is done using an 8-bit quantization with the trained LoRA adaptor. Temperature and \texttt{top\_p} are fixed at 0 and 1 respectively, with a variation of \texttt{maximum\_new\_token} value based on the dataset. ProofWriter requires larger number of tokens at inference when compared to FOLIO because of the size of input. Additionally, as NL-based baseline, separate versions of models are also fine-tuned on textual data without symbolic translations (the corresponding results are reported under Standard). The test data used in our experiments are taken from \citet{pan2023logic} for ProofWriter and ProntoQA. FOLIO comes with a pre-defined dev dataset, that is used for evaluations. 

\subsection{Incremental SFT}
Incremental SFT follows the same process as vanilla SFT for training. The augmented data is passed through LLaMA-2 13B model for fine-tuning. The \texttt{maximum\_new\_token} value is reduced to produce individual generation for incremental inference. This additionally ensures minimal time lag between vanilla and incremental inferences. The incremental setup is performed on FOLIO and ProofWriter datasets. For FOLIO, the whole training set (1000 records) is used and the data augmentation results in an increase in data size ($\sim7000$ records). To measure the generalizability of this technique, we sample 1000 records from our \textsc{ProofFOL} dataset and apply the data augmentation, resulting in $\sim 20000$ records since the number of statements in ProofWriter are larger. We also perform SFT on the original records to create a baseline for these models.

\subsection{Verifier Training}
The verifier model, used during the inference of FOL and predicate generation, is a T5 large model trained on a perturbation dataset. For our experiments, we train a total of 4 verifiers, two for FOLIO and another two for ProofWriter. The perturbations are applied on their respective training data and vary with respect to the complexity of the FOL statements. ProofWriter consists of shorter sentences, that use less operators and predicates, and can have variations of perturbations without excess manual effort. FOLIO, on the other hand, covers a wide range of operators. We run our SFT model on the training dataset to get the perturbations in addition to the defined ones. The proportion of perturbed instances are kept higher than the correct instances in the final verifier training set for all the datasets. The T5 model is trained for 10 epochs with AdamW optimizer~\cite{adamW} and a learning rate of $5 \times e^{-5}$. 

Once trained, the verifiers could run in synchronous (online) or asynchronous (offline) mode. While the online mode corrects errors at each step of  inference, before moving to next step (i.e., correction at time step $t$ impacts step $t+1$ during generation), the offline mode applies corrections on the fully generated predicate and FOLs as a post-processing step (i.e., correction at time step $t$ does not have any consequential effect on $t+1$). For time overhead of incremental decoding and verification, see Table~\ref{tab:incr}.

\section{Results and Discussion}
\begin{table}[t]
    \centering
    \resizebox{1\columnwidth}{!}{%
    \begin{tabular}{p{.4cm}p{0.5cm}cp{2.5cm} c c}
        \toprule
        & \textbf{Type}&& \textbf{Model} & \textbf{Standard} & \textbf{FOL} \\
        \toprule
        \parbox[t]{2mm}{\multirow{8}{*}{\rotatebox[origin=c]{90}{ProofWriter}}} &\multirow{4}{*}{ICL} & \multirow{4}{*}{n-shot} 
        & LLaMA-2 70B & 41.83 & 78.33 \\
        & & & LLaMA-2 13B & 44.16 & 24.66 \\ 
        & & & Mistral 7B & \underline{49.67} & 66.50 \\ 
        & & & Mixtral $8\times7\text{B}$ & 45.83 & \underline{85.00} \\ \cline{2-6}
        &\multirow{4}{*}{SFT}& \multirow{2}{*}{5000} & LLaMA-2 $13$B & 64.16 & 53.50 \\
        & & & Mistral \(7\)B & {70.33} & \underline{\textbf{98.17}} \\ \cline{3-6}
        && \multirow{2}{*}{10000} & LLaMA-2 $13$B & \underline{\textbf{85.66}} & 86.33 \\
        & & & Mistral \(7\)B & 69.50 & 97.83 \\
        \midrule        
        \parbox[t]{2mm}{\multirow{8}{*}{\rotatebox[origin=c]{90}{ProntoQA}}} &\multirow{4}{*}{ICL} & \multirow{4}{*}{n-shot} 
        & LLaMA 70B & 50.60 & \underline{38.80} \\
        & & & LLaMA-2 13B & 47.19 & 11.4 \\ 
        & & & Mistral 7B & 50.60 & 10.80 \\ 
        & & & Mixtral $8\times7\text{B}$ & \underline{58.40} & 13.00 \\ \cline{2-6}
        &\multirow{4}{*}{SFT}& \multirow{2}{*}{5000} & LLaMA-2 $13$B & 55.40 & 53.20 \\
        & & & Mistral $7$B & 60.00 & 78.60 \\ \cline{3-6}
        && \multirow{2}{*}{10000} & LLaMA-2 $13$B & 53.20 & 47.80 \\
        & & & Mistral $7$B & \underline{\textbf{70.40}} & \underline{\textbf{85.40}} \\
                    \midrule    
        \parbox[t]{2mm}{\multirow{8}{*}{\rotatebox[origin=c]{90}{FOLIO}}} &\multirow{4}{*}{ICL} & \multirow{4}{*}{n-shot} 
        & LLaMA-2 70B & 50.74 & 34.97 \\
        & & & LLaMA-2 $13$B & 43.84 & 24.13 \\ 
        & & & Mistral $7$B & 51.23 & {35.96} \\ 
        & & & Mixtral $8\times7\text{B}$ & \underline{57.14} & \underline{\textbf{42.36}} \\ \cline{2-6}
        &\multirow{4}{*}{SFT}& \multirow{2}{*}{5000} & LLaMA-2 $13$B & 40.89 & 26.11 \\
        & & & Mistral $7$B & \underline{\textbf{67.98}} & 26.11 \\ \cline{3-6}
        && \multirow{2}{*}{10000} & LLaMA-2 $13$B & 40.89 & \underline{34.48} \\
        & & & Mistral $7$B & 66.01 & 27.59 \\
        \bottomrule
    \end{tabular}}
    \caption{
        Comparison of models' deductive reasoning accuracy under Standard and FOL-based output prediction. Here ICL denotes ``in-context learning" with n-shots (details of shots in appendix), and SFT denotes ``supervised fine-tuning" (on $5k$ or $10k$ training data subset from \textsc{ProofFOL}). Accuracy metrics in \textbf{bold} signify notably high performance within the same benchmark, while \underline{underline} indicates best results under Standard and FOL for ICL and SFT.}
    \label{tab:finetune}
\end{table}

\subsection{Main Results}\label{sec:main-results}
Table~\ref{tab:finetune} refers to a set of results on logical reasoning dataset using in-context learning and SFT. The main focus of these experiments is to assess how much of a gap exists between large and smaller language models prior to fine-tuning, and how this gap is bridged with further fine-tuning of smaller LMs. LLaMA-2 70B is assumed to be the baseline for LLaMA-2 13B, and Mixtral $8\times7$B model for Mistral 7B. 

\paragraph{Standard} experiment reports free-form reasoning, where the LLM is given a question (premises and a conclusion) and is tasked to produce a direct response. The standard generation varies vastly for n-shot across the datasets. Mistral and Mixtral models show higher accuracy for few-shot learning, but the the results are mixed under SFT. LLaMA-2 $13$B model responses well with the $10k$ examples from \textsc{ProofFOL} for ProofWriter, but does not show a significant gain for FOLIO and ProntoQA. The fine-tuned Mistral models have higher accuracy over the few-shot results for both $5k$ and $10k$ records. This can be attributed to the unreliable output from FOL in the absence of a reasoning path. 

\paragraph{FOL} experiments show correlation of performance with model size, where, for few-shot ICL, LLaMA-2 70B and Mixtral perform consistently better than their smaller versions. ProofWriter achieves 86\% accuracy with LLaMA-2 13B model, fine-tuned on \textsc{ProofFOL}, which is a significant gain in performance when compared to the few-shot setting. Mistral fine-tuned models are the state-of-the-art in FOL generation for ProofWriter datasets, where the model produces 0 syntax errors after fine-tuning. ProntoQA, an altered form of ProofWriter, shows similar trends in performance gain with our fine-tuned models. Mistral 7B fine-tuned on \textsc{ProofFOL} data outperforms all the ProntoQA baselines. For ProntoQA, there is an observed negative effect of overfitting, when LLaMA-2 is trained on larger dataset (increasing training data from $5k$ to $10k$), but we don't notice this for Mistral. We speculate this might be reflective of difference in model size and how it impacts potential training memorization (i.e., overfitting) for larger models. FOLIO, as expected, is a challenging dataset with complex language and structure. Mixtral model shows has the highest accuracy with FOLIO dataset using few-shot at 42\%. LLaMA-2 13b models improve from 24\% to 34\% when fine-tuned, and is on-par with a much larger LLaMA-2 70B. The syntax and semantic error counts for each of these results are specified in the Appendix.

\begin{figure}[t]
    \centering
    \subfigure[]{%
        \includegraphics[trim={0 1cm 0cm 1.1cm},clip,width=\columnwidth]{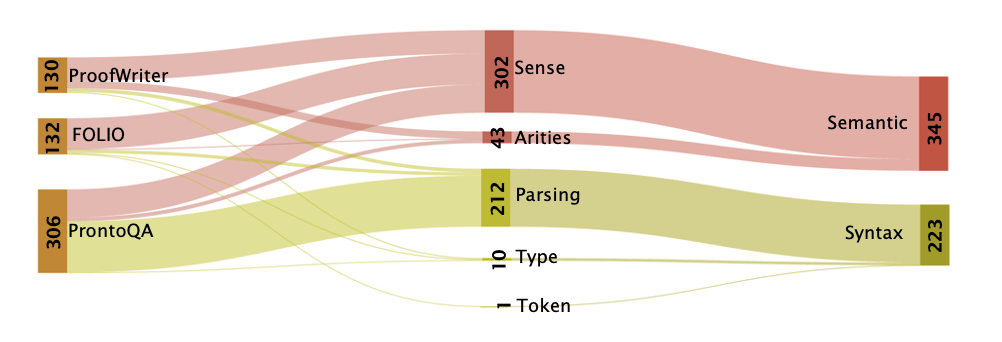}
        \label{fig:subfig1}
    }
    \subfigure[]{%
        \includegraphics[trim={0 1cm 0cm 1.5cm},clip,width=\columnwidth]{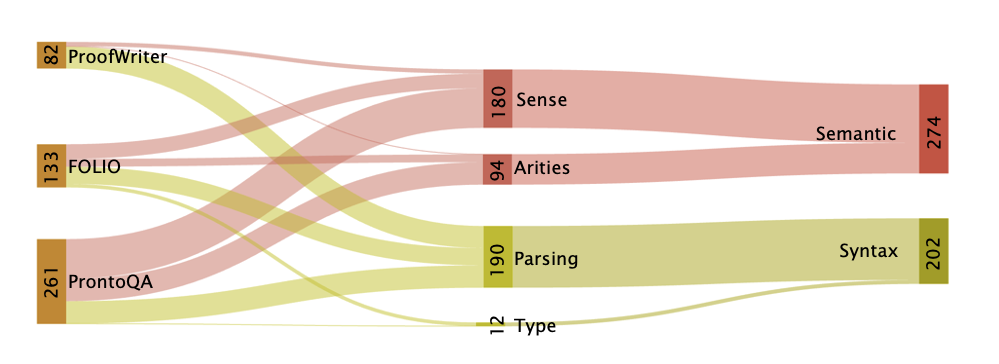}
        \label{fig:subfig2}
    } 
    \caption{Error distribution of LLaMA-2 70B (top) and LLaMA-2 13B (bottom) post fine-tuning on \textsc{ProofFOL}.}
    \label{fig:sanki}
\end{figure}

\paragraph{Error Distribution}
 We provide an error distribution of models on NL-to-FOL translation errors. The results for LLaMA-2 70B with ICL and LLaMA-2 13B after fine-tuning is provided in Fig \ref{fig:sanki}. ProofWriter and ProntoQA show decline in errors, whereas FOLIO stays equivalent to the LLaMA-2 70B ICL model. This plot indicates that a model significantly smaller in size can achieve better generation over a larger model when fine-tuned with relevant data.

\begin{table}[t]
    \centering
    \resizebox{1\columnwidth}{!}{%
    \begin{tabular}{lccc}
        \toprule
        \textbf{Models} & \textbf{Inference} & \textbf{FOLIO}& \textbf{ProofWriter} \\
        \midrule
        ICL Baseline & \textbf{Vanilla} & 24.13 & 24.66 \\\midrule
        SFT Vanilla ($1k)$ & \textbf{Vanilla} & 22.66 (01:55)  & 24.50 (03:36)\\
        SFT Incremental ($1k^\textbf{*})$& \(\textbf{Incremental}\) & 32.02 (02:58) & 27.16 (02:42)\\
        SFT Incremental ($1k^\textbf{*}$) & \(\textbf{+Verifier (On-Off)}\) & \underline{\textbf{37.44}} (03:05) & 29.05 (03:10)\\
        SFT Incremental ($1k^\textbf{*}$) & \(\textbf{+Verifier (On-On)}\) & 29.56 (03:27) & \underline{\textbf{32.50}} (03:31)\\
        \bottomrule
    \end{tabular}}
    \caption{Comparison of LLaMA-2 13B models across FOLIO and ProofWriters under different training and inference protocols. $^\textbf{*}$ the $1k$ vanilla SFT data under incremental SFT expands to $7k$ (FOLIO) and $20k$ (ProofWriter) data points for training. The inference time for SFT models for each record is reported on an average in the format (MM:SS). \textbf{Vanilla}: All predicates and FOLs are generated in one pass. 
    \textbf{Incremental}: All predicates are generated in one pass and then fed into the next step along with premises and conclusions to generate FOLs one by one. \textbf{+Verifier (On-Off)}: Predicate verifiers are called during inference (Online) while FOL verifiers are called once all FOLs are generated (Offline). \textbf{+Verifier (On-On)}: Applies both verifiers during inference (both in online mode).}
    \label{tab:incr}
\end{table}

\subsection{Incremental Results}
Incremental techniques are performed on ground truth FOLIO and a subset of \textsc{ProofFOL} dataset. The intention behind the incremental technique is to show that the data augmentation method improves the translations over the original dataset. In Table~\ref{tab:incr}, we focus on the shift in performance between LLaMA-2 13B model fine-tuned on the original 1000 records and augmented $1000\times n$ records, where n represents the scale at which the data grows after augmentation. Both FOLIO and ProofWriter have low performance in FOL generation when trained with a small dataset. FOLIO shows a steady improvement using vanilla and incremental inferences. This pushes us to use a verifier to further the performance. With the predicates corrected during generation and FOLs corrected after inference, this verifier based incremental setting achieves 37\% using Offline verifier, which outperforms LLaMA-2 70B ICL accuracy 34.97\% (Table~\ref{tab:finetune}). ProofWriter, when trained incrementally, provides varied results with different inferences. The verifier inference model achieves 29\% an 32\% accuracy when compared to the SFT model on original dataset, 24\%. It can be noted that the performance of the FOLIO model is lower when the FOL verifier is online, since any error from the verifier is passed on to the next generation and can cause a domino effect. The change in inference times between incremental and verifier settings is minimal. \footnote{Since ProofWriter has larger number of sentences, we use an adaptive inference token size, where token with less than 5 words have lower token size. For example, ``Dog chases the cat'' can be translated to ``Chases(Dog, Cat)'', which require less than 16 tokens. This adaptive technique lowers incremental time for ProofWriter.}

\begin{table}[htbp]
    \centering
    \begin{tabular}{ccc}
        \toprule
        \textbf{Instruction} & \textbf{Accuracy}& \textbf{Syntax errors} \\
        \midrule
                               \(\text{Mixed}\) & 35.46 & 64\\
                               \(\text{Single}\) & 35.47 & 70\\
                               \(\text{Ordered}\) & 32.02 & 63\\
                               \(\text{Unique}\) & 21.18 & 80\\
                               \(\text{Check}\) & 27.09 & 108\\
        \bottomrule
    \end{tabular}
    \caption{Comparison of Incremental Methods with  Instruction Fine-tuning. Each method represents the type of instruction used for fine-tuning and inference. The syntax errors are out of 203 test records in FOLIO dataset.}
    \label{tab:abl}
\end{table}

\subsection{Ablation Studies}
The incremental methods follows certain rules of finetuning and generation. Our model uses two instructions for predicate and FOL generation. We performed an ablation study on using different variation of incremental training and inference for FOLIO dataset. The type of ablations and their performances are given in Table~\ref{tab:abl}. These were performed before training the verifier module. In place of the verifier, we initially used Prover9 to check for syntax and any invalid generation followed a sampling by the LLM, and the first error free FOL was selected. This method proved ineffective as the sampling method was time-consuming. The results in the table are after the tool verification, except for the \emph{Check} model, where we use LLM as a verifier. 

\emph{Mixed} is the current method of incremental training and inference, where we use different instructions for predicate and FOL generation. \emph{Single} method uses only one instruction;`Complete the generation'. This performs equivalent to the \emph{Mixed}, but results in additional syntax errors, presumably because of the vague instruction. \emph{Ordered} uses different instructions for each FOL generation. The instructions carry information about the sentence that is required to be translated. This method helped keep the syntax errors low, but lowered the overall performance. Instead of passing the previously generated FOL to the next generation, we applied \emph{Unique}, where the statements are split and passed one at a time with the predicate values. This performs poorly and does not include passage level translations. In \emph{Check}, we use the perturbation dataset with specific instructions along with the training dataset for fine-tuning. We instruct the model to identify and correct the errors at inference. This method relies on the language model to perform a new task with limited data and proves to be ineffective. 

\section{Related Work}
\paragraph{LLM for symbolic translation}
The use of formal language translations by LLMs was initially attempted by \citet{nye2021improving}, with an intent to emphasize the importance of dual process theory for logical reasoning tasks. Following this, the process of reasoning was offloaded to theorem provers and LLMs served as systems to generate symbolic translations \cite{pan2023logic, ye2024satlm, olausson2023linc}. The available research in this method majorly differs in variation of formal language (used by different theorem provers) and verification process to handle translation errors. These methods make use of the expensive and ambiguous\footnote{We refer to ambiguity in the data used to train the GPT model.} GPT models, restricting the symbolic framework to non-critical domains. Corresponding to the work in formal logic, \citet{yang2023harnessing} applied supervised finetuning to LLaMA model to improve the natural language to first-order logic translations at a sentence level. Our research shifts the focus to building a complete translation system that can handle multiple statements. In addition to this, we perform a systematic analysis of translation errors which enabled us to build a verification mechanism.

\paragraph{Symbolic Decoding with LLMs}
The choice of decoding strategies can improve text generation, specifically in LLMs where the output follows a structured format. In neuro-symbolic models, neuro-logic decoding \cite{lu2020neurologic, lu2021neurologic} applies symbolic constraints. Interactive theorem provers were also used alongside LLMs to ensure a constrained generation of the reasoning path \cite{poesia2023certified}. Other techniques like contrastive step-wise decoding helped with improving the probability of a correct reasoning path \cite{su2023llms}. The success of these symbolic decoding strategies motivates our research to apply verification during the inference stage. 

\paragraph{Deductive reasoning benchmarks}
Deductive reasoning requires logical derivation of conclusion using a set of premises. The available benchmarks, ProofWriter \cite{tafjord2020proofwriter} and ProntoQA \cite{prontoqa}, show a reasoning path for identifying the validity of question from its context. These datasets have a simple, yet multi-hop deductive style. We develop FOL data for the training set of ProofWriter dataset to make it suitable for translation tasks.  FOLIO \cite{han2022folio} is another deductive task which is semantically complex with human annotated FOL sentences. The presence of gold-standard FOL should ideally make FOLIO suitable for developing or improving formal language translation models, however the limited size of this dataset impedes the realization of this goal. We present data augmentation technique to overcome this issue. To the best of our knowledge, our work is the first at using the incremental setting, with augmentation and verification, in the context of logical reasoning with NL. 

\section{Conclusion}
Formal language translation systems for logical reasoning task worked effectively in the era of LLMs, with persisting translation errors. In this paper, we provided an understanding of general translation errors by LLMs, when used as NL-to-FOL translation systems. We highlighted the importance of first-order logic (FOL) ground truth data and present a pipeline to generate high quality FOLs for ProofWriter dataset, introducing an FOL-annotated data called \textsc{ProofFOL}. Using \textsc{ProofFOL}, we fine-tuned a set of smaller language models and showed an increase in performance over larger LMs. Additionally, the issue of data scarcity is addressed via proposing incremental techniques, which cover data augmentation, inference verification, and correction. Our experiments on 3 benchmarks highlight the potential of our proposed framework.

\newpage
\bibliography{aaai25}

\newpage
\appendix
\newpage 
\section{Data Generation}
\begin{figure}[hbt]
  \centering
  \includegraphics[width=\columnwidth]{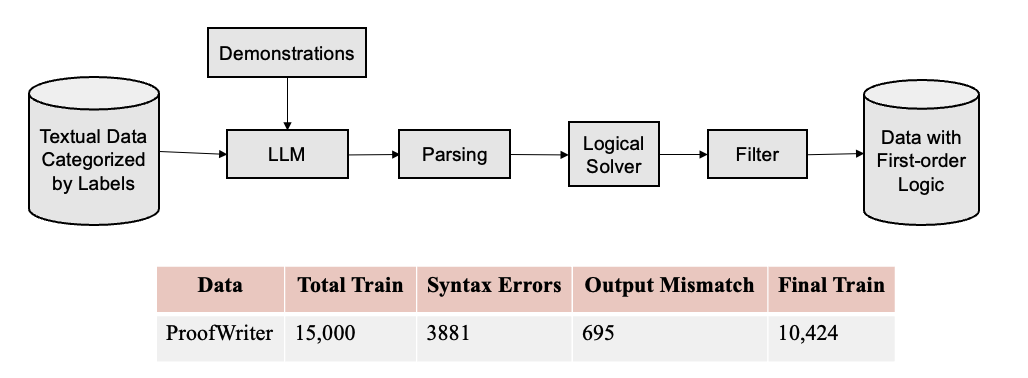}
  \caption{Overview of the Data Generation Pipeline and Key Statistics}
  \label{fig:gen}
\end{figure}

The data pipeline applied for generating ProofFOL is detailed in Fig \ref{fig:gen}. The train data is 15000 data points with depth-5 from ProofWriter dataset. The 15k records are sampled randomly ensuring a fair distribution of the labels; True, False, and Uncertain. The LLM here is GPT4o with a output token length of 1000 for each generation. We use $url=/v1/chat/completions$ format for batch generations of GPT4o. The logical solver is Prover 9, a theorem prover suitable to run in python environment with nltk library, that uses CNF conversions, quantifier operations, and skolemization to transform the clauses into a tree format. Parsing errors by Prover 9 occur when the FOL formula cannot be converted to a tree structure because it does not follow specific grammar rules. After filter adn parsing stages, we get 10424 records with FOL statements. Syntax errors are the errors thrown by the tool and semantic errors are measured by comparing the ground truth label with the solver output.

\begin{figure}[hbt]
  \centering
  \includegraphics[width=\columnwidth]{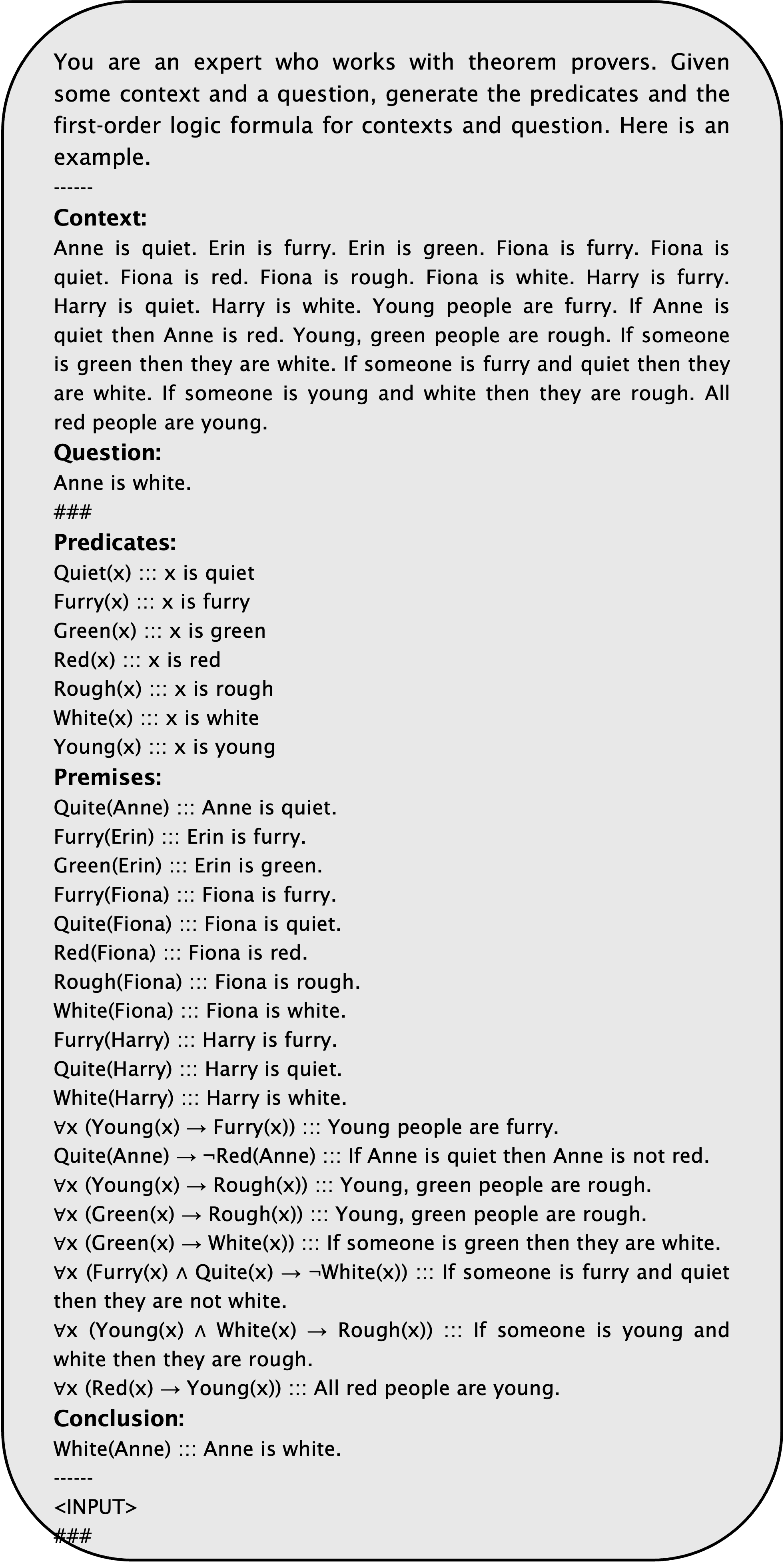}
  \caption{Few shot example with instruction for Data Generation process}
  \label{fig:fsdatagen}
\end{figure}

\begin{figure}[t]
  \centering
  \includegraphics[width=\columnwidth]{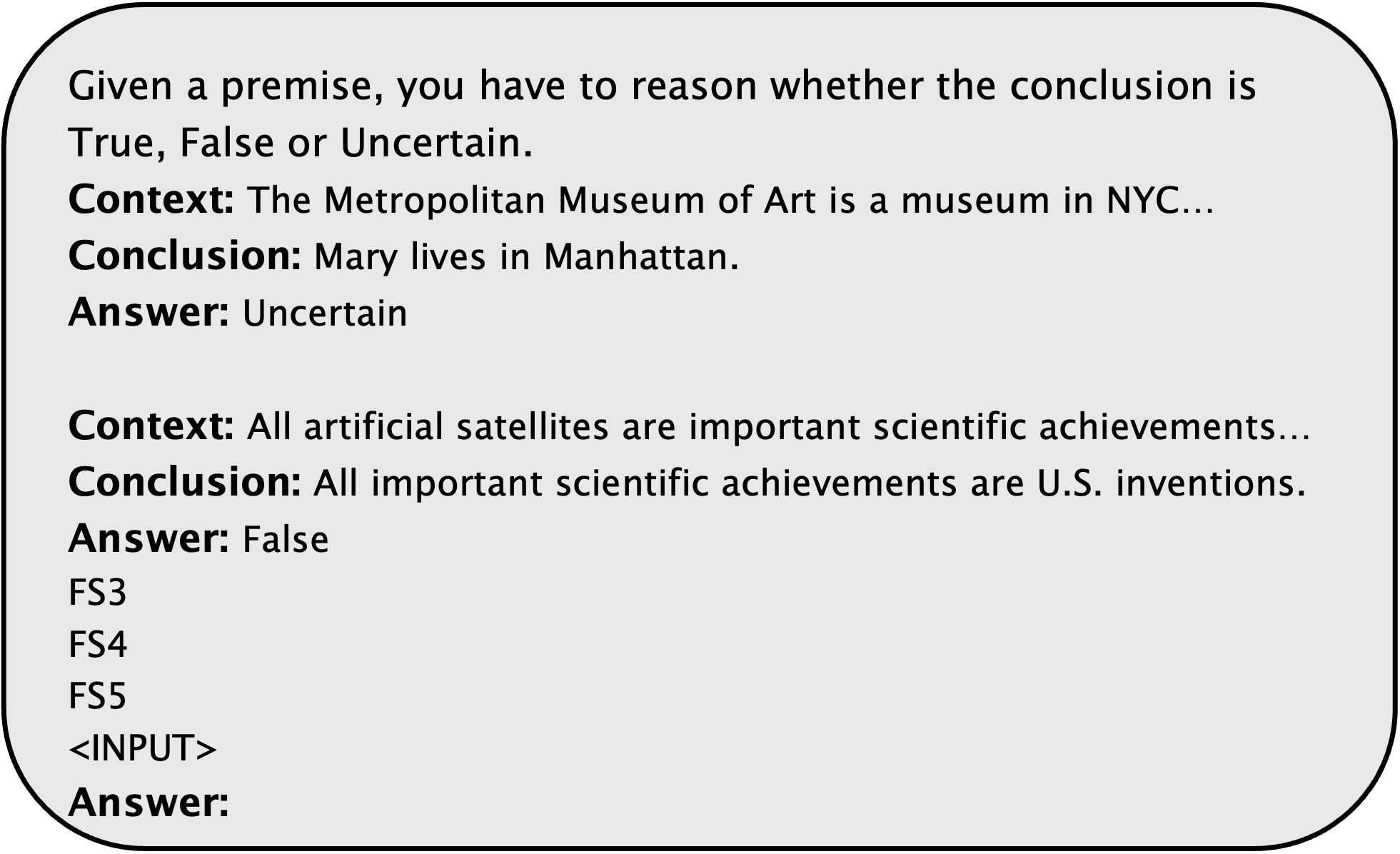}
  \caption{Standard Generation Few-shot examples for FOLIO: For brevity, we highlighted the format and the instructions and minimized the content in the few-shot examples.}
  \label{fig:fsstanfolio}
\end{figure}

\begin{figure}[t]
  \centering
  \includegraphics[width=\columnwidth]{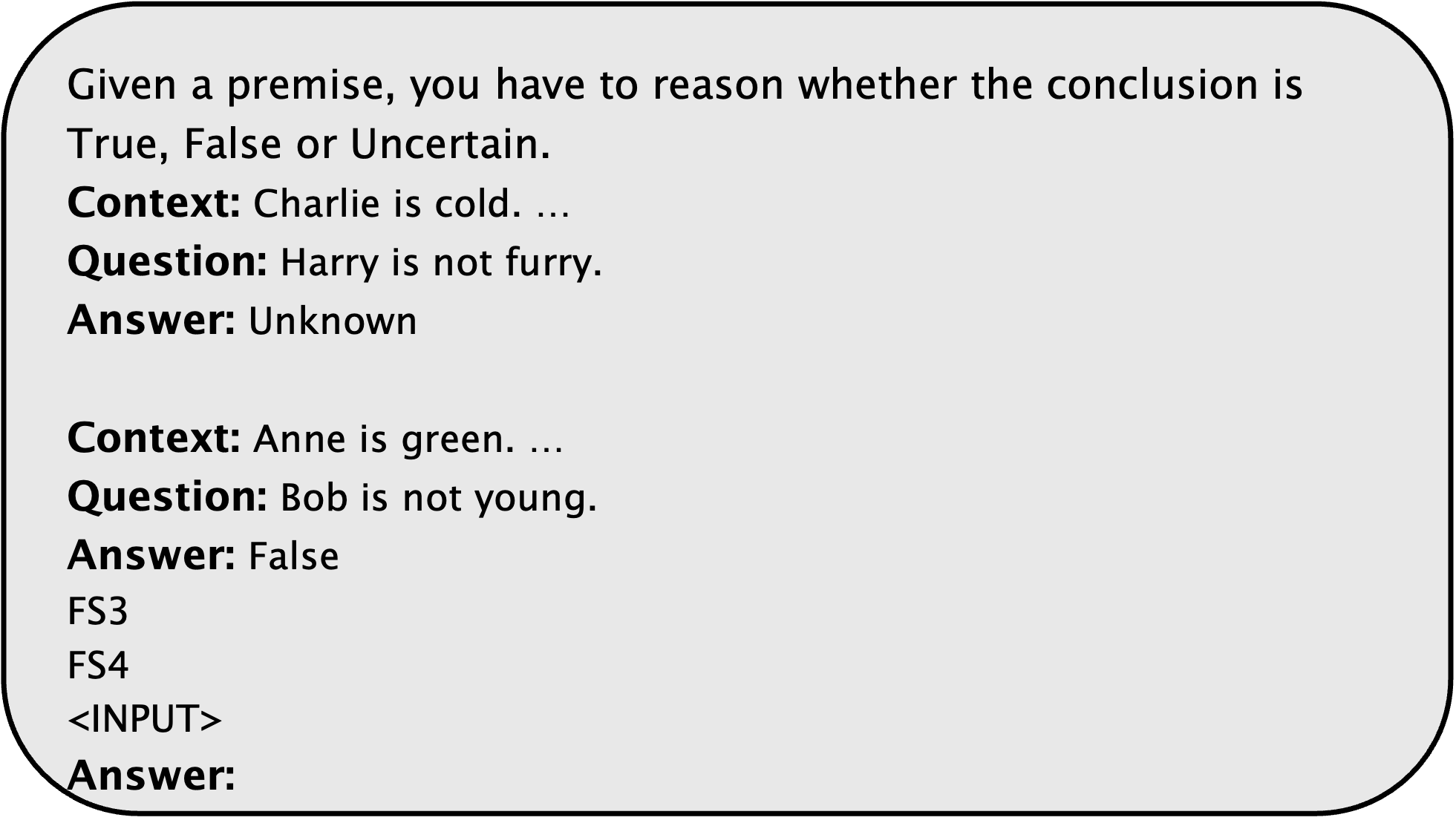}
  \caption{Standard Generation Few-shot examples for ProofWriter and ProntoQA}
  \label{fig:fsstanproof}
\end{figure}

\begin{figure}[t]
  \centering
  \includegraphics[width=\columnwidth]{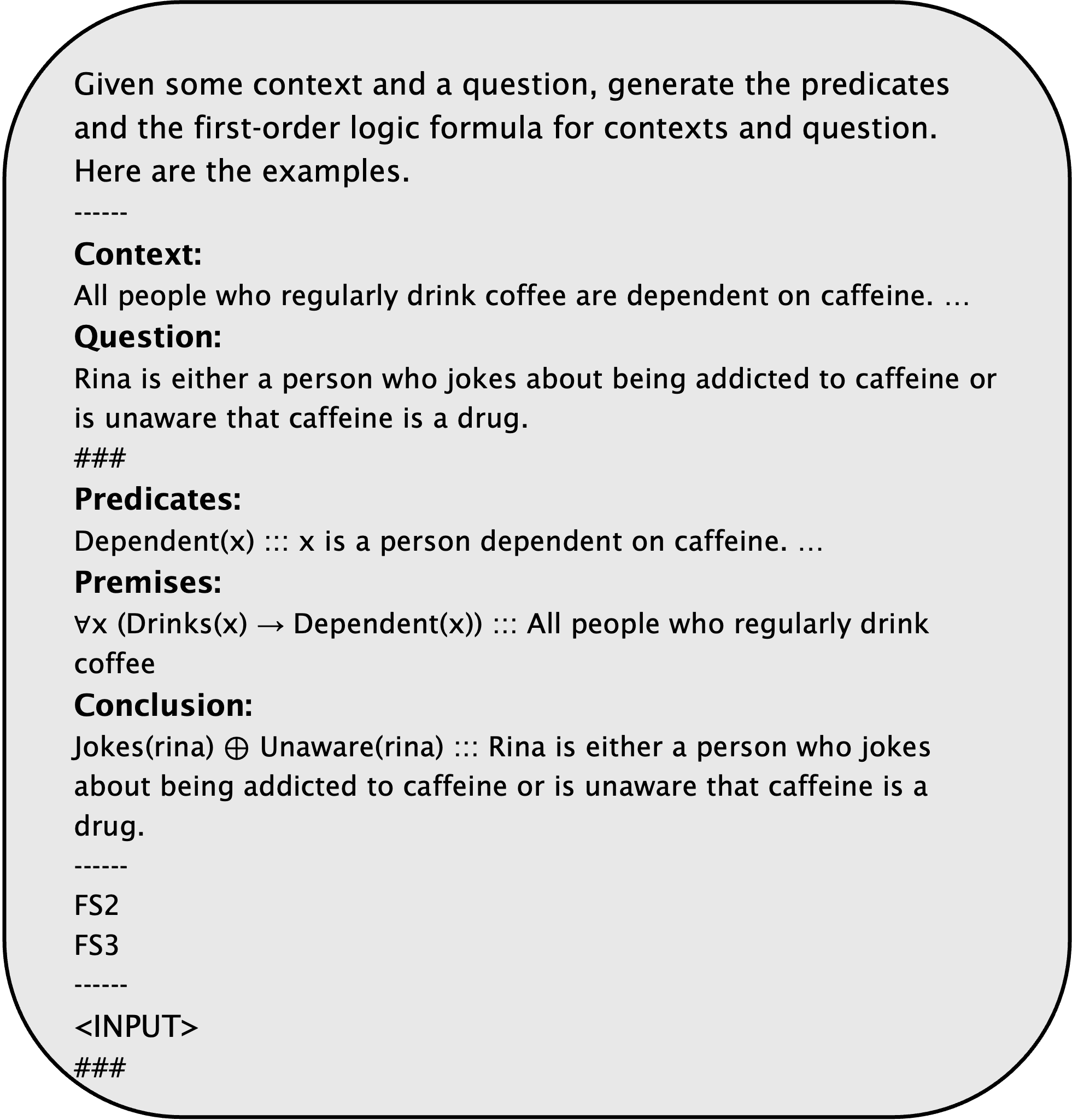}
  \caption{FOL Generation Few-shot examples for FOLIO}
  \label{fig:fsfolfolio}
\end{figure}

\begin{figure}[t]
  \centering
  \includegraphics[width=\columnwidth]{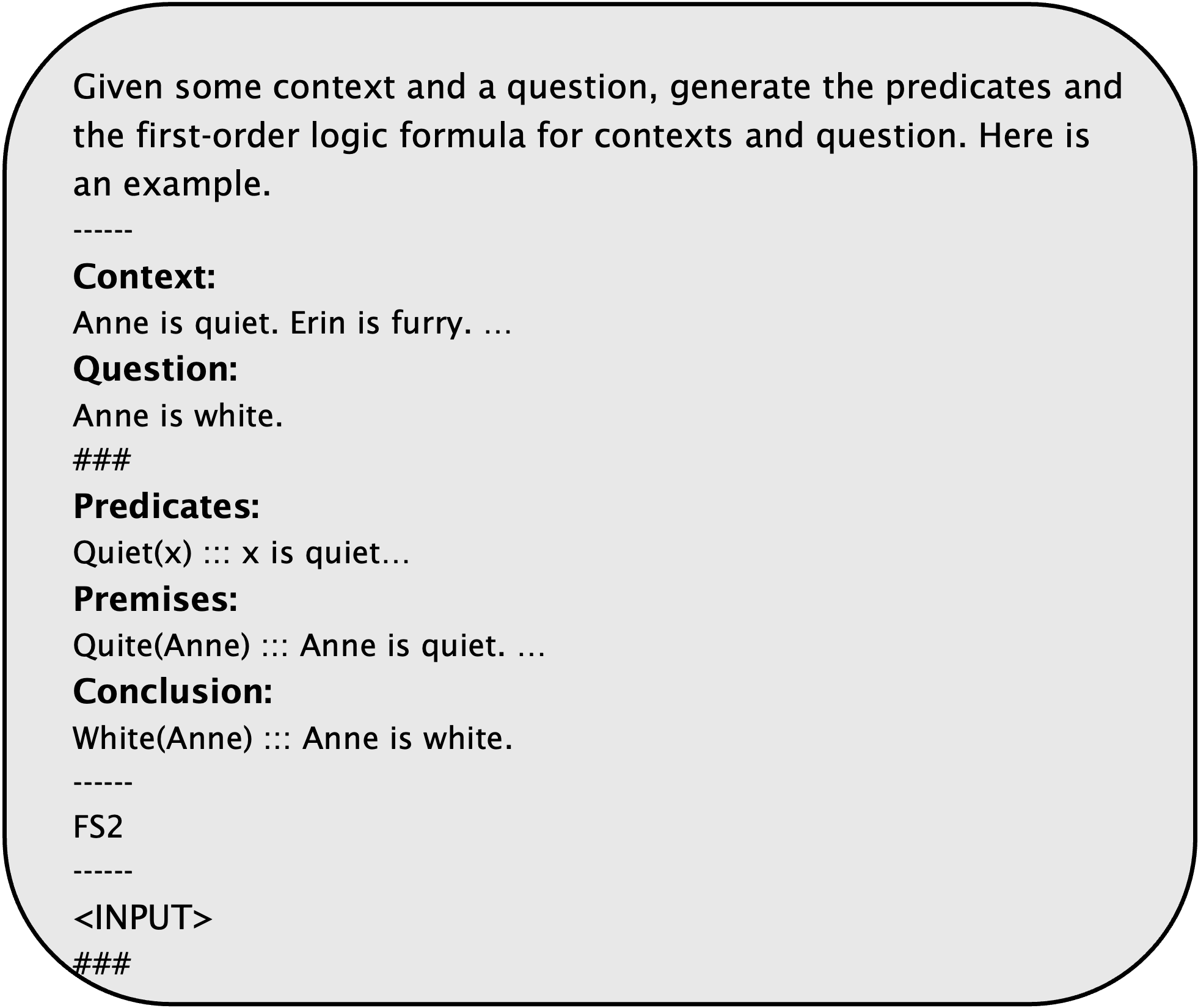}
  \caption{FOL Generation Few-shot examples for ProofWriter and ProntoQA}
  \label{fig:fsfolproof}
\end{figure}

\section{Few Shot Examples}
In this section, the few-shot format used for all the in-context learning tasks are presented. 
\paragraph{Few-shot for Data-augmentation}
The Fig \ref{fig:fsdatagen} shows the format of the few-shot example for data generation. We tried this with 2 examples for 50 unique training data points, but it did not generate better translation results. Instead, we ensured that all operators are covered in this example, specifically negation. We first generate Predicates associated with the sentence. This is followed by FOL generation. Each FOL generation comes with the natural language and this format is adapted from (Pan et al. 2023). We use the same format of generation for all of our experiments with an exception of Predicates description, as this part is harder to verify. The description is removed from ProofFOL.

\begin{itemize}
    \item Standard generation: For standard generation, we randomly sample examples from the training set. We use 5-shot for FOLIO and 4-shot for ProofWriter and ProntoQA as shown in Fig \ref{fig:fsstanfolio} and Fig \ref{fig:fsstanproof}
    \item FOL generation: For FOL generation, we use partially use examples from (Pan et al. 2023) and sample the rest from the training set. The FOL syntax is fixed to represent Prover9 format. We use 3-shot for FOLIO and 2-shot for ProofWriter and ProntoQA as shown in Fig \ref{fig:fsfolfolio} and Fig \ref{fig:fsfolproof}

\end{itemize}

\begin{table}[htbp]
    \centering
    \begin{tabular}{c|c|c}
        \toprule
        \textbf{Dataset} & \textbf{Standard}& \textbf{FOL} \\
        \midrule \hline
                               \(\text{ProofWriter}\) & 4 & 2\\
                               \(\text{FOLIO}\) & 5 & 3\\
                               \(\text{ProntoQA}\) & 4 & 2\\
        \bottomrule
    \end{tabular}
    \caption{Number of Few-shot examples used for creating baselines in Table 1}
    \label{tab:fs}
\end{table}

\section{Datasets}
We use three datasets for our experiments. 
\begin{itemize}
    \item ProofWriter: The training set for ProofWriter is sampled from the depth-5 records. For test-set, we use the one provided in (Pan et al. 2023)
    \item ProntoQA: ProntoQA is a test dataset. We use ProofWriter examples as training set and test-set sample from (Pan et al. 2023)
    \item FOLIO: FOLIO has 1001 training set records and 203 dev set. We use the original training set for SFT models and dev for evaluating the model.
\end{itemize}

\section{Data Augmentation}
\begin{figure}[hbt]
  \centering
  \includegraphics[width=\columnwidth]{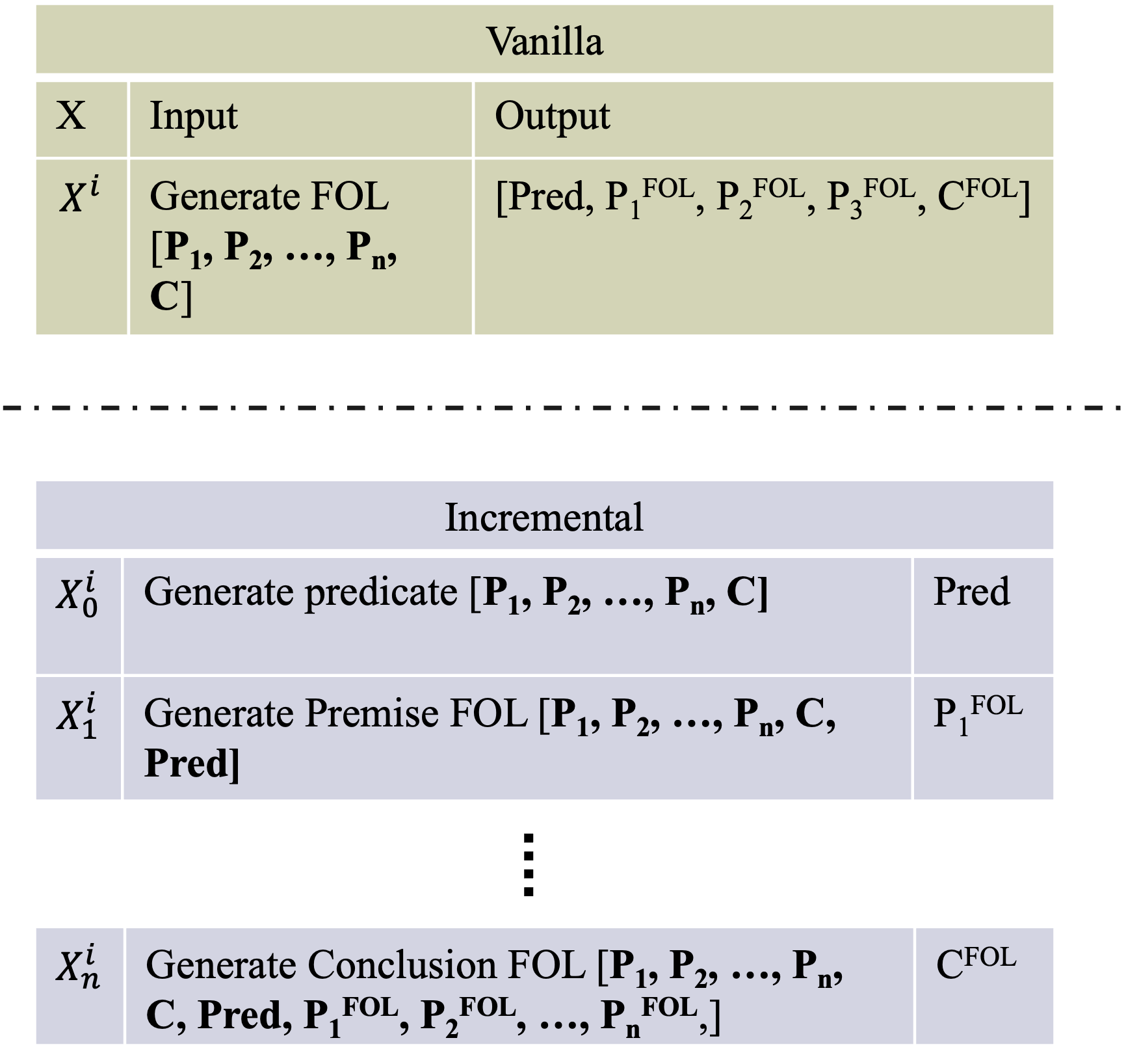}
  \caption{Data augmentation: the vanilla represents original data point. Incremental is the data that is present after augmentation.}
  \label{fig:dataaug}
\end{figure}

Data augmentation for FOL generation uses incremental data assignment, where the output is divided into multiple tasks, as shown in Fig \ref{fig:dataaug}. This method is applied on two datasets; ProofWriter and FOLIO. A subset of the ProofWriter dataset is extracted from the ProofFOL data and passed for data augmentation, increasing the size to 20X the original. For FOLIO, we take the full dataset and perform augmentation, making it 7X larger. Given the style of incremental data, we remove few records from FOLIO dataset which do not follow the one-to-one mapping of text and FOL. The size stats are detailed in Table~\ref{tab:dataaug}. 

\begin{table}[htbp]
    \centering
    \begin{tabular}{c|c|c}
        \toprule
        \textbf{Dataset} & \textbf{Original}& \textbf{Augmented} \\
        \midrule \hline
                               \(\text{ProofWriter}\) & 1000 & 7288 \\
                               \(\text{FOLIO}\) & 998 & 20145\\
        \bottomrule
    \end{tabular}
    \caption{Size of training datasets before and after data augmentation}
    \label{tab:dataaug}
\end{table}

\section{SFT Models}
\begin{table}[t]
    \centering
    \resizebox{1\columnwidth}{!}{%
    \begin{tabular}{p{.4cm}p{0.5cm}cp{2.5cm} c c}
        \toprule
        & \textbf{Type}&& \textbf{Model} & \textbf{Syntax} & \textbf{Semantic} \\
        \toprule
        \parbox[t]{2mm}{\multirow{8}{*}{\rotatebox[origin=c]{90}{ProofWriter(600)}}} &\multirow{4}{*}{ICL} & \multirow{4}{*}{n-shot} 
        & LLaMA-2 70B & 44 & 86 \\
        & & & LLaMA-2 13B & 314 & 138 \\ 
        & & & Mistral 7B & 130 & 71 \\ 
        & & & Mixtral $8\times7\text{B}$ & 44 & 46 \\ \cline{2-6}
        &\multirow{4}{*}{SFT}& \multirow{2}{*}{5000} & LLaMA-2 $13$B & 71 & 15 \\
        & & & Mistral \(7\)B & 0 & 11 \\ \cline{3-6}
        && \multirow{2}{*}{10000} & LLaMA-2 $13$B & 69 & 13 \\
        & & & Mistral \(7\)B & 0 & 13 \\
        \midrule        
        \parbox[t]{2mm}{\multirow{8}{*}{\rotatebox[origin=c]{90}{ProntoQA(500)}}} &\multirow{4}{*}{ICL} & \multirow{4}{*}{n-shot} 
        & LLaMA 70B & 203 & 103 \\
        & & & LLaMA-2 13B & 266 & 177 \\ 
        & & & Mistral 7B & 420 & 26 \\ 
        & & & Mixtral $8\times7\text{B}$ & 413 & 22 \\ \cline{2-6}
        &\multirow{4}{*}{SFT}& \multirow{2}{*}{5000} & LLaMA-2 $13$B & 108 & 126 \\
        & & & Mistral $7$B & 26 & 81\\ \cline{3-6}
        && \multirow{2}{*}{10000} & LLaMA-2 $13$B & 139 & 122 \\
        & & & Mistral $7$B & 13 & 60 \\
                    \midrule    
        \parbox[t]{2mm}{\multirow{8}{*}{\rotatebox[origin=c]{90}{FOLIO(203)}}} &\multirow{4}{*}{ICL} & \multirow{4}{*}{n-shot} 
        & LLaMA-2 70B & 19 & 113 \\
        & & & LLaMA-2 $13$B & 95 & 59 \\ 
        & & & Mistral $7$B & 95 & 35 \\ 
        & & & Mixtral $8\times7\text{B}$ & 79 & 38 \\ \cline{2-6}
        &\multirow{4}{*}{SFT}& \multirow{2}{*}{5000} & LLaMA-2 $13$B & 103 & 47 \\
        & & & Mistral $7$B & 64 & 86 \\ \cline{3-6}
        && \multirow{2}{*}{10000} & LLaMA-2 $13$B & 88 & 45 \\
        & & & Mistral $7$B & 48 & 99 \\
        \bottomrule
    \end{tabular}}
    \caption{
        Syntax and Semantic error count for FOL generation}
    \label{tab:app_error}
\end{table}

Table~\ref{tab:app_error} is an extension of Table~\ref{tab:incr} from the main paper. It shows how the syntax and semantic errors vary with ICl and fine-tuning. There is a constant trend of lower syntax errors with fine-tuning the models with relevant data for both Llama and Mistral models.

\section{Error Analysis}
The NL-FOL translation errors can be identified using the tool feedback or a mismatch in tool output with the ground truth. We identify these issues and categorize them into syntax and semantic errors. Fig \ref{tab: syntax} shows examples of each type of error. These are identified by manually analysing the FOL statements in cases where there was no feedback from the tool. The details of each error are provided here.
\begin{itemize}
    \item Missing quantifier: When a predicate includes a variable, it must have either a Universal Quantifier ‘$\forall$’ or an Existential Quantifier ‘$\exists$’. This error occurs if either quantifier is missing.
    \item Parenthesis error: The formula becomes invalid if there is an extra or a missing parenthesis.
    \item Completion error: The FOL is either incomplete or contains additional text that disrupts its logical flow.
    \item Quantifier location: Quantifiers that are either repetitive or incorrectly positioned result in grammatical inaccuracies in the expression.
    \item Missing variable: When multiple quantifiers are present, the tool fails to parse predicates that lack free variables.
    \item Special token: The tool does not handle special characters in the input.
    \item Unknown operator: The tool does not support parsing mathematical equations.
    \item Predicate error: These errors arise when predicates are reused with different subjects, omitted entirely, or conjoined inappropriately, leading to erroneous interpretations of the logical constructs in FOL statements. Such misinterpretations can affect the accuracy and reliability of responses generated by LLMs.
    \item Incorrect quantifier: This occurs when an existential quantifier is incorrectly chosen in situations that require a universal quantifier, leading to a logical contradiction between the terms. This mismatch between the chosen quantifier and the necessary logical condition can result in flawed reasoning and inconsistencies in logical analysis.
    \item Predicate mismatch: It occurs when LLMs are tasked with generating predicates based on text passages and fail to recognize synonymous terms as equivalent. This results in the tool counting synonyms as distinct tokens, leading to discrepancies in predicate generation.
    \item Arities error: It occurs when predicates are inconsistently applied with a varying number of constants across different statements. Such discrepancies can introduce ambiguity in logical inferences. This is a semantic error that is captured by the tool.
    \item Subject predicate: The logic is flawed when a subject is used both as a predicate and a constant in the expression.
\end{itemize}
\begin{table}[h]
    \centering
    \begin{tabular}{p{0.3cm}p{1.5cm}|>{\raggedright\arraybackslash}p{5cm}}
    \toprule
    \textbf{} & \textbf{Type} & \textbf{Example} \\
    \cline{2-3}
     \multirow{9}{*}{\makebox[0pt][c]{\rotatebox[origin=c]{90}{Parsing}}} &Missing Quantifier & $\text{BerkeleyCollege}(\underline{x})$ $\land \text{ResidentialCollegeAt}(\underline{x},$ \text{yaleUniversity}) \\
    \cline{2-3}
    &Parenthesis error  & $\text{BeneficialTo}(\text{cherry}, \text{people})$ $\oplus \text{On}(\text{cherry}, \text{warningList})\underline{)}$ $\to \neg \text{RedFruit}(\text{cherry})$ \\
    \cline{2-3}
    &Completion error  & $\forall x (\text{Athlete}(x)$ $\to$$ \neg \text{NeverExercises}(x))$ \underline{\text{Never: does not exist a time}} \\
    \midrule
    \multirow{7}{*}{\makebox[0pt][c]{\rotatebox[origin=c]{90}{Type}}}&Quantifier location & $\underline{\exists y} (\text{Own}(\text{emily}, y)$ $\land \text{Roommate}(y)) \to$ $\underline{\exists y} (\text{Own}(\text{emily}, y)$ $\land \text{LiveIn}(\text{emily}, \text{apartment}))$ \\
    \cline{2-3}
    &Missing variable  & $\forall x \exists y (\text{\underline{In(indonesia)}} \land \text{Prosecutor}(x) \land \text{SpecialCrime}(y) \rightarrow \text{InvestigatePersonally}(x, y))$ \\ 
    \midrule
    \multirow{4}{*}{\makebox[0pt][c]{\rotatebox[origin=c]{90}{Token}}}& Special $ $    token  & Endowment(yale, \underline{42.3} billion) \\ 
    \cline{2-3}
    &Unknown Operator & \begin{tabular}[c]{@{}l@{}}$\forall x (\text{Rating}(x, y) \land y \underline{>} 4$$ \rightarrow \text{Listed}(x))$\end{tabular} \\ 
    \midrule
    \midrule
     \multirow{7}{*}{\makebox[0pt][c]{\rotatebox[origin=c]{90}{Sense}}}&Predicate errors & Error: \underline{$\neg$Solid2Pointers}(jack) $\land$ \underline{Successful3Pointers}(jack)
    
    True: $\neg$GoodAt(jack, twos) $\land$ GoodAt(jack, threes) \\
    \cline{2-3}
    &Incorrect Quantifier  & Error: \underline{$\exists x$} (FleaBeetle(x) $\rightarrow$ $\neg$InFamily(x, chrysomelidae))

    True: $\forall x$ (FleaBeetle(x) $\rightarrow$ $\neg$In(x, chrysomelidaeFamily)) \\
    \cline{2-3}
    &Predicate Mismatch & Error: $\neg$High(NewHaven); \underline{Low(towerA)}
    
    True: $\neg$High(NewHaven); $\neg$High(towerA) \\
    \midrule
     \multirow{9}{*}{\makebox[0pt][c]{\rotatebox[origin=c]{90}{Arities}}}&Arities    errors & Error: Sees(Tiger, Mouse); $\forall x$(((Visits(x, Rabbit)) $\land$ \underline{(Sees(Mouse)))} $\rightarrow$ (Visits(x, Tiger)))
    
    True: Sees(Tiger, Mouse); $\forall x$ (((Visits(x, Rabbit)) $\land$ (Sees(x, Mouse))) $\rightarrow$ (Visits(x, Tiger))) \\
    \cline{2-3}
    &Subject Predicate  & $\underline{\text{Platypus}(\text{platypus})}$ $\land \neg \text{Teeth}(\text{platypus})$ $\land \text{Mammal}(\text{platypus})$ \\
    \bottomrule
    \end{tabular}
    \caption{Common Errors in First-Order Logic: The first block of errors are syntactic and the second are semantic errors. This table categorizes errors by their cause, with the \textit{\underline{underlined}} text highlighting the specific cause or location of each error. For semantic errors, there are 'True' values to make sense of the error. }
    \label{tab: syntax}
\end{table}

\section{Verification Perturbations}
The perturbations used for training T5 verifier models are designed from the errors in the previously discussed error analysis. The perturbations are different for FOLIO and ProofWriter dataset, as FOLIO is a complex dataset with additional operators when compared to ProofWriter. To handle the complexity of FOLIO dataset, we pass the training set data to the SFT model and match the translations with ground truth. Any mismatch is treated as a perturbation.  Other perturbations are based manually included. This dataset has a portion of correct values, making it a verification and correction system. The count of these datasets is detailed in Table~\ref{tab:pertstats}.
\begin{table}[htbp]
    \centering
    \begin{tabular}{c|c|c|c|c}
        \toprule
        \textbf{Dataset} & \textbf{Verifier}& \textbf{Correct}& \textbf{Training}& \textbf{Manual} \\
        \midrule \hline
                               \(\text{ProofWriter}\) &  Predicate & 1978 & 326& 2991 \\
                               \(\text{ProofWriter}\) & FOL & 2375 & 1595& 2358\\
                               \(\text{FOLIO}\) & Predicate & 1724 & -& 3448\\
                               \(\text{FOLIO}\) & FOL & 2000 & -& 3241\\
        \bottomrule
    \end{tabular}
    \caption{Amount of data for each type of perturbations}
    \label{tab:pertstats}
\end{table}

\paragraph{FOLIO Perturbations}
The predicate perturbations are designed based on the commonly observed errors in the predicates. These are not complex errors, but usually a missing predicate or variable. Based on this, we use three types of perturbations.
\begin{itemize}
    \item Omit One Predicate: We randomly omit a predicate from all predicates.
    \item Omit One Variable: We choose a predicate with the maximum number of variables and omit the last one in the chosen predicate.
    \item Omit Both Variable and Predicate: We choose a predicate with the maximum number of variables, omit the last argument in the chosen predicate, and also randomly omit a predicate from the rest of the predicates.
\end{itemize}

The FOL perturbations tackle the syntax errors that are common while generating FOL for FOLIO test data.
\begin{itemize}
    \item Change Quantifier Position: We move the quantifier position to different positions in a FOL statement.
    \item Omit One Quantifier: We randomly omit one quantifier from a FOL statement.
    \item Omit Last Bracket: We omit the last bracket in a FOL statement.
\end{itemize}

\paragraph{ProofWriter Perturbations}
The ProofWriter dataset has simpler, but larger number of predicates. based on this, we design 4 types of perturbations.

\begin{itemize}
    \item Omit one predicate: We randomly omit a predicate from all predicates.
    \item Omit or add one variable: We choose a random predicate, and if the predicate consists of multiple variables, we omit one, and if it consists of only one variable, we add one.
    \item Add plural predicates: To handle the synonymous predicate issue, we randomly select a predicate, create a plural form, and add it to the set of predicates.
    \item Duplicate predicate: We select a random predicate and add a variable to it. This is added back to the set of predicates.
\end{itemize}

In addition to the quantifier, FOL generations in ProofWriter consists of three major operators; and, imply, and negation. We include variations in this for our perturbation data. ProofWriter can be divided into simple FOL statements (facts) and complex ones (rules). The simple statements usually do not contain any operators or quantifiers. Based on this, we design 5 types of perturbations.
\begin{itemize}
    \item Add or omit negations: We randomly select facts and either add a negation or remove an existing one.
    \item Omit arguments from facts: We randomly select predicates from facts and omit the variables if the predicate has more than one variable.
    \item Add or omit quantifier: We randomly select rules and either add a quantifier or remove an existing one.
    \item Swap operators: We replace `and' with `imply' operator or vice-versa if the operator exists in the FOL statements.
    \item Add plural predicates: To handle the synonymous predicate issue, we randomly select a predicate, create a plural form, and add it to the set of predicates.
\end{itemize}

\paragraph{Incremental and Verification Ablation}
We present additional ablation done on the incremental methods and analyse their results. 
\begin{itemize}
    \item Incremental Ablation: The incremental finetuning uses two instruction; generate predicates and generate FOL (brief). This allows us to perform a full scale inference rather than an incremental one, by increasing the output token size. We first let the model predict the predicates and the input along with the predicates is passed to the LLM for full FOL generation. Since the data in the augmented set consist of such a case (last value in augmentation), the model is able to generate the FOL in a flow. We performed inference in this manner and the results are in  Table~\ref{tab:incr-appendix}, where vanilla* represents the full FOL inference. The results in ProofWriter are higher than the incremental results, but on closer observation, we determine that the model hallucinates few cases in the vanilla* and helps with better performance. The incremental method allows a strict format of generation and forces the model to generate the FOL for one statement at a time, without adding additional values. Additionally, because of this mismatch in number of statements, we cannot use a FOL verifier on vanilla*.
\end{itemize}

\begin{table}[t]
    \centering
    \resizebox{1\columnwidth}{!}{%
    \begin{tabular}{>{\raggedright\arraybackslash}p{2.2cm}|>{\raggedright\arraybackslash}p{1.9cm}|p{1.5cm}}
        \toprule
        \textbf{Inference} & \textbf{FOLIO}& \textbf{ProofWriter} \\
        \midrule
                                 vanilla* & 31.52 & \underline{64.00}\\
                                 \hline
                                \(\text{Incremental}\) & 32.02 & 27.16\\
                                \(\text{+Verifier}\) & \underline{37.44} & 29.05\\
        \bottomrule
    \end{tabular}}
    \caption{Comparing two forms of inference for incremental models. This inference ablation is done on FOLIO and ProofWriter datasets.}
    \label{tab:incr-appendix}
\end{table}

\section{System Requirements for Experimentation}
The Llama 2 model weights are downloaded from the official Llama website \url{https://llama.meta.com/llama-downloads/}. Mistral-7B (\url{https://huggingface.co/mistralai/Mistral-7B-v0.3}) model and Mixtral-8x7B(\url{https://huggingface.co/mistralai/Mixtral-8x7B-Instruct-v0.1}) are accessed via the Hugging Face interface. All the models are gated and require access, which is typically a safety measure and can be easily granted. The transformer version used in our experiments is $v4.40.0$ and should ideally be above $v4.28.0$ or the latest version to run Llama and Mistral models without any errors. All the other library requirements will be specified in the code. 

\end{document}